%% file: emnlp2023.tex
\pdfoutput=1

\documentclass[11pt]{article}

\usepackage{EMNLP2023}

\usepackage{times}
\usepackage{latexsym}
\usepackage{listings}
\usepackage{makecell}
\usepackage{booktabs}
\usepackage[T1]{fontenc}

\usepackage[utf8]{inputenc}

\usepackage{microtype}

\usepackage{inconsolata}

\usepackage{graphicx}
\usepackage{subfigure}
\usepackage[symbol]{footmisc}

\title{GraphextQA: A Benchmark for Evaluating Graph-Enhanced Large Language Models}

\author{
Yuanchun Shen$^{\footnotemark[1]\;\;3}$, \; Ruotong Liao$^{\footnotemark[1]\;\;1,2}$, \; Zhen Han$^{\footnotemark[2]\;\;5}$, \; Yunpu Ma$^{1,4}$, \;  Volker Tresp$^{1,2}$ \\
$^{1}$LMU Munich $\;$ $\;$  $^{2}$Munich Center for Machine Learning (MCML) $\;$ \\
$^{3}$Technical University of Munich $\;$  $^{4}$Siemens AG  $\;$  $^{5}$Amazon\\ 
\texttt{y.c.shen@tum.de, ruotong.liao@outlook.com,\; } \\
\texttt{cognitive.yunpu@gmail.com, hanzhen02111@gmail.com, volker.tresp@lmu.de}
}

\begin{document}

\maketitle

\addtocounter{footnote}{1}
\footnotetext{Equal contribution.}
\addtocounter{footnote}{1}
\footnotetext{Work done prior to joining Amazon.}

\begin{abstract}

While multi-modal models have successfully integrated information from image, video, and audio modalities, integrating graph modality into large language models (LLMs) remains unexplored. This discrepancy largely stems from the inherent divergence between structured graph data and unstructured text data. Incorporating graph knowledge provides a reliable source of information, enabling potential solutions to address issues in text generation, e.g., hallucination, and lack of domain knowledge. To evaluate the integration of graph knowledge into language models, a dedicated dataset is needed. However, there is currently no benchmark dataset specifically designed for multimodal graph-language models. To address this gap, we propose GraphextQA\footnote{The dataset is available at \url{https://huggingface.co/datasets/drt/graphext-qa}}, a question answering dataset with paired subgraphs, retrieved from Wikidata, to facilitate the evaluation and future development of graph-language models. Additionally, we introduce a baseline model called \textit{CrossGNN}\footnote{The model is available at \url{https://github.com/happen2me/cross-gnn}}, which conditions answer generation on the paired graphs by cross-attending question-aware graph features at decoding. The proposed dataset is designed to evaluate graph-language models' ability to understand graphs and make use of it for answer generation. We perform experiments with language-only models and the proposed graph-language model to validate the usefulness of the paired graphs and to demonstrate the difficulty of the task.

\end{abstract}

\input{1_intro}

\input{2_related}

\input{3_dataset}

\input{4_model}

\input{5_exp}

\input{6_concl}

\input{7_limit}

\input{8_ethic}

\input{9_appendix}

\bibliography{anthology,custom}
\bibliographystyle{acl_natbib}

\end{document}

%% file: 1_intro.tex
\section{Introduction}

Multi-modal models, such as visual-language and audio-language models, have shown impressive capabilities in integrating information from various modalities into large language models (LLMs). However, the integration of the graph modality into LLMs remains relatively unexplored. Integrating graphs into LLMs offers an additional trustworthy source of knowledge and extends the model's ability to comprehend this widely existing modality. It may also facilitate an easier understanding of graph information for users by explaining the encoded information in natural language. 

Currently, the evaluation of cross-modal integration from graphs to LLMs lacks dedicated tasks and datasets. A related task in this context is information retrieval-based knowledge base question answering (KBQA), where natural language questions are answered by predicting the appropriate nodes in relevant subgraphs retrieved from knowledge graphs \cite{ijcai2021kbqasuervey}. These relevant subgraphs shed light on the possible approaches to evaluate graph-language models --- by assessing the improvement achieved through the integration of these subgraphs, it is possible to evaluate a language model's ability to understand graph information. Nevertheless, the presence of useful information within these graphs is not guaranteed  \cite{sun-etal-2019-pullnet,yasunaga2022dragon,zhang-etal-2022-subgraphretrieve}, making them unsuitable for the direct evaluation of graph-language models. For example, it becomes challenging to determine whether issues arise from uninformative graphs or the model's inability to comprehend the graph modality when an LLM fails to answer a question.

To bridge this gap, we introduce Graph-text Question Answering (GraphextQA), an open-domain question answering dataset that includes paired graphs for developing and evaluating graph-language models. The open-domain questions necessitate a deep understanding of real-world knowledge. This knowledge is conveniently provided in the form of graphs within the dataset. The graphs are sourced from Wikidata and consist of reasoning paths from entities mentioned in the questions to the entities that the questions ask. The objective of this dataset is to assess the LLM's ability to leverage graph information. It also facilitates the development of algorithms that integrate knowledge from graphs into language models. 

As there are no existing LLMs specifically designed for graph understanding, we also introduce a baseline model called CrossGNN to bridge this gap and to show the difficulty of the proposed task. CrossGNN builds upon a frozen T5 model, and conditions the answer generation with question-aware graph features encoded with a graph neural network (GNN). CrossGNN serves as a foundation for exploring the intersection of graph understanding and generative language models.


\begin{figure*}[t]
 \centering
  \subfigure{\includegraphics[width=0.4\textwidth]{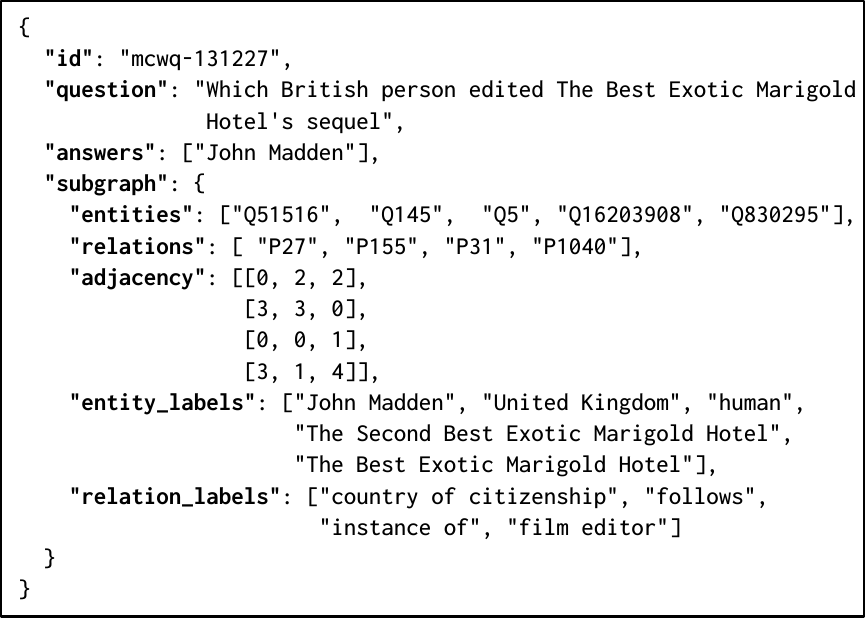}}
  \hspace{1.5cm}
  \subfigure{\includegraphics[width=0.25\textwidth]{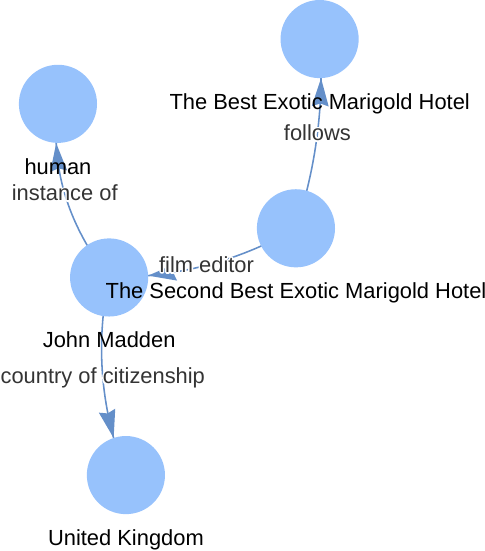}}
  \caption{Left: Example from GraphextQA dataset. Right: Visualization of the corresponding graph. The graph is represented as an adjacency list in GraphextQA, where each fact is stored as [subject index, predicate index, object index]. The subject and object indices correspond to their positions in the local entity list, while the predicate index corresponds to its position in the local relation list. The labels of the relations and entities are included to aid human in understanding the graph.}
  \label{fig:data-example}
\end{figure*}

%% file: 2_related.tex
\section{Related Works}


\subsection{Existing datasets in KBQA.}

Existing datasets in Knowledge Base Question Answering (KBQA) can be categorized into two types based on whether logical forms are provided. The first type is designed for semantic parsing-based (SP-based) methods, where questions are parsed into logical forms and executed against a knowledge graph to obtain answers \cite{cui-etal-2022-compositional,perevalov2022qald9}. These datasets provide both the questions and their corresponding logical forms. To answer these questions, models usually use sequence-to-sequence models to translate the natural language questions to graph queries. The second type of dataset is designed for information retrieval-based methods. These approaches construct question-specific subgraphs from large knowledge graphs and rank entities within the subgraph to obtain the answer entities. Such datasets usually provide only questions and answer entities \cite{longpre-etal-2021-mkqa,sen-etal-2022-mintaka}. However, neither of these two datasets provides pertinent and precise paired subgraphs. Moreover, most of existing KBQA datasets, such as WebQuestions\cite{berant2013semantic}, ComplexQuestions\cite{bao-etal-2016-constraint}, WebQuestionSP\cite{yih-etal-2016-webqsp}, ComplexWeb questions\cite{talmor-berant-2018-cwq}, and Grailed QA\cite{gu2021beyond}, are based on Freebase \cite{bollacker2008freebase}, a knowledge graph that ceased updating in 2015. A few are designed for up-to-date knowledge graphs, such as KQA pro \cite{cao-etal-2022-kqa}, Lc-QuAD 2.0 \cite{dubey2019lcquad2}, and MCWQ \cite{cui-etal-2022-compositional}. Among them, logical forms from KQA Pro are not executable on Wikidata. Therefore, we mainly base our dataset on Lc-QuAD 2.0 and MCWQ.

\subsection{Knowledge Graph Embeddings}

Similar to the initialization of language tokens with pretrained token embeddings in the language modality, pretrained knowledge graph embeddings (KGE) can be used to initialize knowledge graph entities and relations. These embeddings capture the structure and semantic information of the knowledge graph by representing entities and relations as continuous vectors in a vector space \cite{wang2017kgesurvey}. Various algorithms, such as TransE, DistMult, ComplEx, RotatE can be employed to train these knowledge graph embeddings. The trained KGE models have proven effective in tasks such as link prediction and relation extraction. In CrossGNN, we leverage the pretrained KGE from Graphvite \cite{zhu2019graphvite} to initialize the node embedding.

\subsection{Integration of Graph into Language Models}

Researchers have explored various approaches to integrating knowledge graph information into language models.

One approach is to enhance language representations with knowledge graphs during pretraining. Models such as KnowBert \cite{peters-etal-2019-knowbert}, EaE \cite{fevry-etal-2020-eae}, ERNIE-THU \cite{zhang-etal-2019-ernie}, and DRAGON \cite{yasunaga2022dragon} incorporate graph information into language models by leveraging entity embeddings, entity memory layers, fusion layers, and cross-modal encoders. These models aim to encode both text and graph information simultaneously, but their primary focus is on encoding rather than language generation, limiting their suitability for generative tasks.

Another approach involves converting knowledge graph triples into text and incorporating them as model inputs. This bridging of the gap between the graph and text modalities explicitly converts knowledge graph triples into textual representations \cite{li2023graph-reason-triplet,agarwal-etal-2021-tkgen}, without the understanding of the graph modality.


There have also been attempts to integrate knowledge graph embeddings into language generation. For example, \citep{zhou2018commonsensegen} retrieves relevant knowledge subgraphs based on user posts and generates responses by attentively reading the retrieved knowledge graphs.  ConceptFlow \cite{zhang-etal-2020-grounded} incorporates graph embeddings through graph neural networks (GNNs) into context representation and utilizes them to predict the next word distribution, including both vocabulary-based words and entity label words. These approaches focus on cross-modality understanding but do not fully exploit the generative capabilities of the models, as they rely on selecting nodes from the knowledge graph as generated texts.

%% file: 3_dataset.tex
\section{GraphextQA: A Graph Understanding Dataset for Language Models}

\subsection{The Task}

Using the GraphextQA dataset, our objective is to evaluate the ability of generative models to comprehend graphs and generate accurate answers. Each instance in the dataset consists of a natural language question and a corresponding graph that represents the necessary reasoning path from mentioned entities to the answers. The model is tasked with reading the question, interpreting the information encoded in the graph, and generating the appropriate answers.

\subsection{Input and Output}

Figure \ref{fig:data-example} illustrates an example from the GraphextQA dataset, which includes a question and a corresponding graph as input. The questions in GraphextQA primarily seek factual information that can be answered using entity labels from Wikidata. The graphs are represented as collections of (subject, predicate, object) triple patterns, outlining the logical steps for answering the question. The output in GraphextQA consists of a list of potential answers, with most questions having a single answer. It is worth noting that the desired model output is, however, a natural sentence that contain one or a combination of multiple answers.



\subsection{Source Datasets}
\label{subsec:source-dataset}

GraphextQA is derived from two complex semantic parsing-based KBQA datasets on Wikidata: Lc-QuAD 2.0 \cite{dubey2019lcquad2} and MCWQ \cite{cui-etal-2022-compositional}. These two datasets ask models to parse questions into SPARQL queries that are executable on Wikidata endpoints \cite{vrandevcic2014wikidata} that retrieve answers. For example, the parsed SPARQL query for the question \textit{Which British person edited The Best Exotic Marigold Hotel's sequel?} from MCWQ dataset is as follows:

\begin{lstlisting}
SELECT DISTINCT ?x0 WHERE {
    ?x1 wdt:P155 wd:Q830295 .
    ?x1 wdt:P1040 ?x0 .
    ?x0 wdt:P27 wd:Q145 }
\end{lstlisting}

The SPARQL query captures three requirements of the question in the \texttt{WHERE} clause. Firstly, the sequel (\textit{?x1}) follows (\textit{P155}) the book \textit{The Best Exotic Marigold Hotel} (\textit{Q830295}); secondly, the sequel is written by (\textit{P1040}) the asked author (\textit{?x0}); thirdly, the asked author has citizenship (\textit{P27}) of the United Kingdom (\textit{Q145}).

\subsection{Dataset Creation}

One of the distinctive features of GraphextQA compared to previous datasets is the inclusion of paired graphs, which serve as additional graph modality knowledge for graph-language models. We argue that a useful graph for answer generation is one that contains a reasoning path from the known information in the question (such as mentioned entities and relations) to the answers. Fortunately, such graphs can be automatically retrieved from the SPARQL queries in semantic parsing-based KBQA datasets.

\subsubsection{Graph Creation}

\begin{figure*}
  \centering
  \includegraphics[width=.8\textwidth]{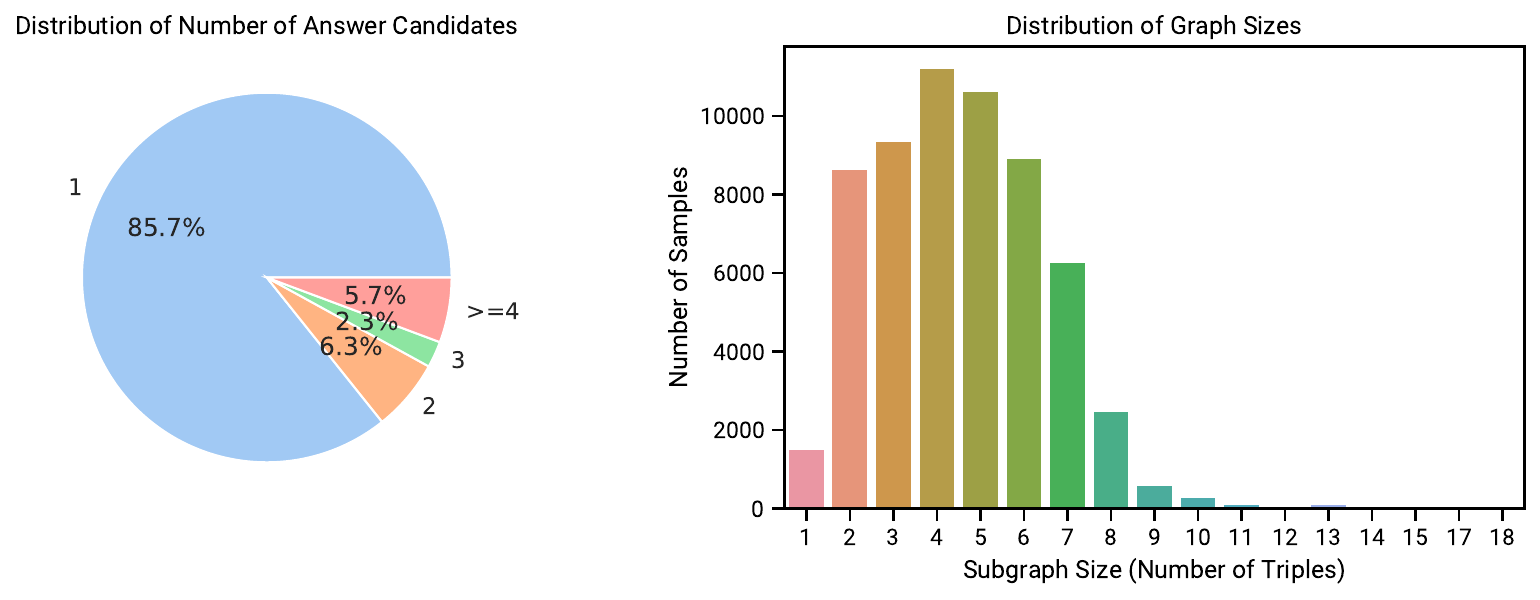}
  \caption{Statistics of the number of answers and triple patterns per question in GraphextQA. The majority of questions have a single possible answer, while others may have multiple answers. Graph size is measured by the number of triple patterns. GraphextQA provides compact paired graph containing relevant information.}
  \label{fig:statistics}
\end{figure*}

To construct the paired graphs, we extract the triple patterns from the WHERE clause of the SPARQL queries. The variables in these patterns are replaced with the queried entities or relations obtained from Wikidata endpoints. Consider the example from section \ref{subsec:source-dataset}. By substituting the variables \texttt{x0} and \texttt{x1}, the three triples in the WHERE clause form a graph that connects known information from the question to the answer.


To retrieve all the unknown variables, we modify the existing SPARQL queries in the KBQA datasets. The intermediate entities are left out from the queried results in existing KBQA datasets, as the queries aim to retrieve the answer entities. For example, only \texttt{x0} is retrieved, but \texttt{x1} is left out in the previous example. However, the intermediate entities are indispensable as the reasoning chain will be incomplete without them. We replace the \texttt{SELECT ?var} command with \texttt{SELECT *} to retrieve every variable in the basic graph patterns. 

Next, we run the queries against a local Wikidata endpoint, leveraging a container provided by \citet{willerval2022qendpoint}, to speed up query and avoid unnecessary strain on public resources. The knowledge base used in this service was created from a truthy snapshot \footnote{This version of dump limits the included statements to direct, truthy ones: \url{https://www.wikidata.org/wiki/Wikidata:Database_download\#Database_dumps}} in May 2021.


Finally, the graph is created by substituting the variables in the triple patterns from the WHERE clauses. The labels of the entities and the relations are also stored for interpretation purposes. Besides, the graph is stored in the format of edge list, where the local entities and local relations are stored in two lists, and the edges are stored as a list of triples like \textit{[subject index, predicate index, object index]}. An example is shown in \ref{fig:data-example}.  Notably, any FILTER clauses within the WHERE clauses were disregarded during the graph construction process.

\subsubsection{Answer Generation}

For MCWQ dataset, we directly leverage the answers from the original dataset and answers. For the Lc-QuAD 2.0 dataset, however, the answer is not included. We instead use the labels of the querying results of the original paired queries as answers.

\subsubsection{Data Selection}

Several considerations were taken into account during the design of the dataset. Firstly, yes or no questions are excluded from GraphextQA dataset. There exists yes or no questions in MCWQ and Lc-QuAD 2.0 datasets, where the question is verified by examining whether the constraints from the WHERE can be satisfied. Such SPARQL starts with the keyword \texttt{ASK}. The triples patterns in the WHERE clause are identical to those in SELECT queries. If the triple pattern does not exist in Wikidata, it returns false. Therefore, for questions with no answer, we are not able to retrieve any graphs. As keeping only yes questions makes answering them trivial, we exclude such questions. Secondly, samples that we fail to construct a graph are also excluded. This can result from the update of the knowledge graph or the miss of information so that the query does not return anything from the dataset. Thirdly, samples with ill-formed answers are excluded. This includes samples with no answers, or those whose answers can not be retrieved because natural language outputs are expected.

\subsection{Dataset Statistics}

GraphextQA consists of 59,964 paired questions, answers, and graphs. Among them, 86.7\% of all questions (52,015) is derived from the MCWQ dataset, while the rest are from Lc-QuAD 2.0 dataset. On average, each graph contains 4.54 triples, each question has 1.5 answers, and each answer has a span of 2.5 words (separated by space). The distribution of the number of answer candidates and graph sizes is shown in Figure. Moreover, the dataset covers 41,255 different entities and 492 different relations from Wikidata. Notably, at least one of the answers is already covered in the paired graph in 97.8\% of all samples, ensuring the relevance of the paired graph.

\begin{figure}
  \centering
  \includegraphics[width=0.3\textwidth]{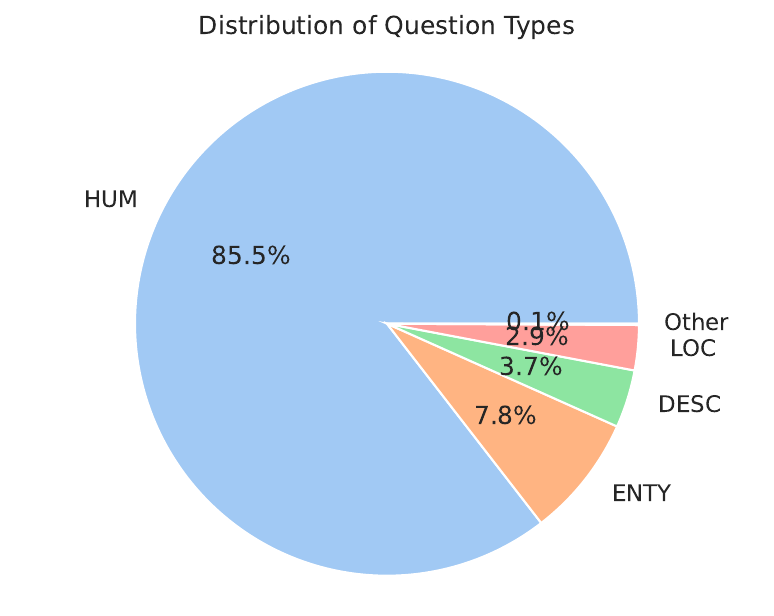}
  \caption{The question type distribution classified with a TREC(50) \cite{voorhees1999trec} question classifier. HUM questions ask to find out a human individual;  ENTY questions ask about general entities like an organization, a religion, or a framework. DESC questions ask about description and abstract concepts, like a physical phenomenon, a strategy, etc. LOC questions ask about a location, like a city, a state, etc.}
  \label{fig:types}
\end{figure}

We employed TREC50 \footnote{\url{https://sparknlp.org/2020/05/03/classifierdl_use_trec50_en.html}} to classify the question into 50 subcategories. We aggregate different subcategories into their main category. For example \textit{DESC\_def} and \textit{DESC\_manner} that ask to describe the definition of something and to ask the manner of an action are aggregated to the main category \textit{DESC}. The results are shown in Figure \ref{fig:types}. It shows that most of the questions ask about human, yet a small portion asks about general entities, descriptions, or locations. Notably, this classifier is not accurate. For example, some questions classified as ENTY actually ask about humans. Yet it gives a rough distribution of what kind of questions are there in the dataset.

\subsection{Metrics}

Existing KBQA methods usually adopt  accuracy, recall, F1, and Hits@1 \cite{yasunaga2022dragon,li2023graph-reason-triplet,ijcai2021kbqasuervey} as predicts entities, which are primarily designed for evaluating classification or ranking way. However, they are not suitable for open-ended generative methods, as the output is not limited to a predefined set of options.  As GraphextQA is designed for generative models, we instead adopt exact match (EM), F1, and BLEU \cite{papineni-etal-2002-bleu} to evaluate generated answers. Notably, the f1 here is based on the token level instead of the option level as in traditional KBQA systems. For situations where there are multiple answers, we will preprocess the generated answers by separating each answer by common connection words before calculating the exact match and f1 score.

%% file: 4_model.tex
\section{Baseline Model}

In this section, we introduce CrossGNN, a graph language baseline model that accepts texts and graphs as input and generates corresponding texts in response to the input text with the information from the input graph. The model builds upon a transformer-like \cite{vaswani2017attention} encoder-decoder model, i.e. T5, for both its ability to encode input texts and its ability to generate free-form responses. The model is built on a fully frozen pre-trained T5 model with extra modules or layers such that it preserves generation ability, avoids catastrophic forgetfulness, and can condition the output on the graph at the same time. During training, only the extra layers are trainable. On the encoder side, we add a text-aware graph encoder to align the graph closer to language modality and extract semantic-relevant knowledge from the graph. On the decoder side, we insert gated cross-attention layers inside language decoding blocks to allow the decoder selectively integrate the extra knowledge from the graph modality.

\subsection{Question-aware Graph Encoder}

\begin{figure*}
  \centering
  \includegraphics[width=.8\textwidth]{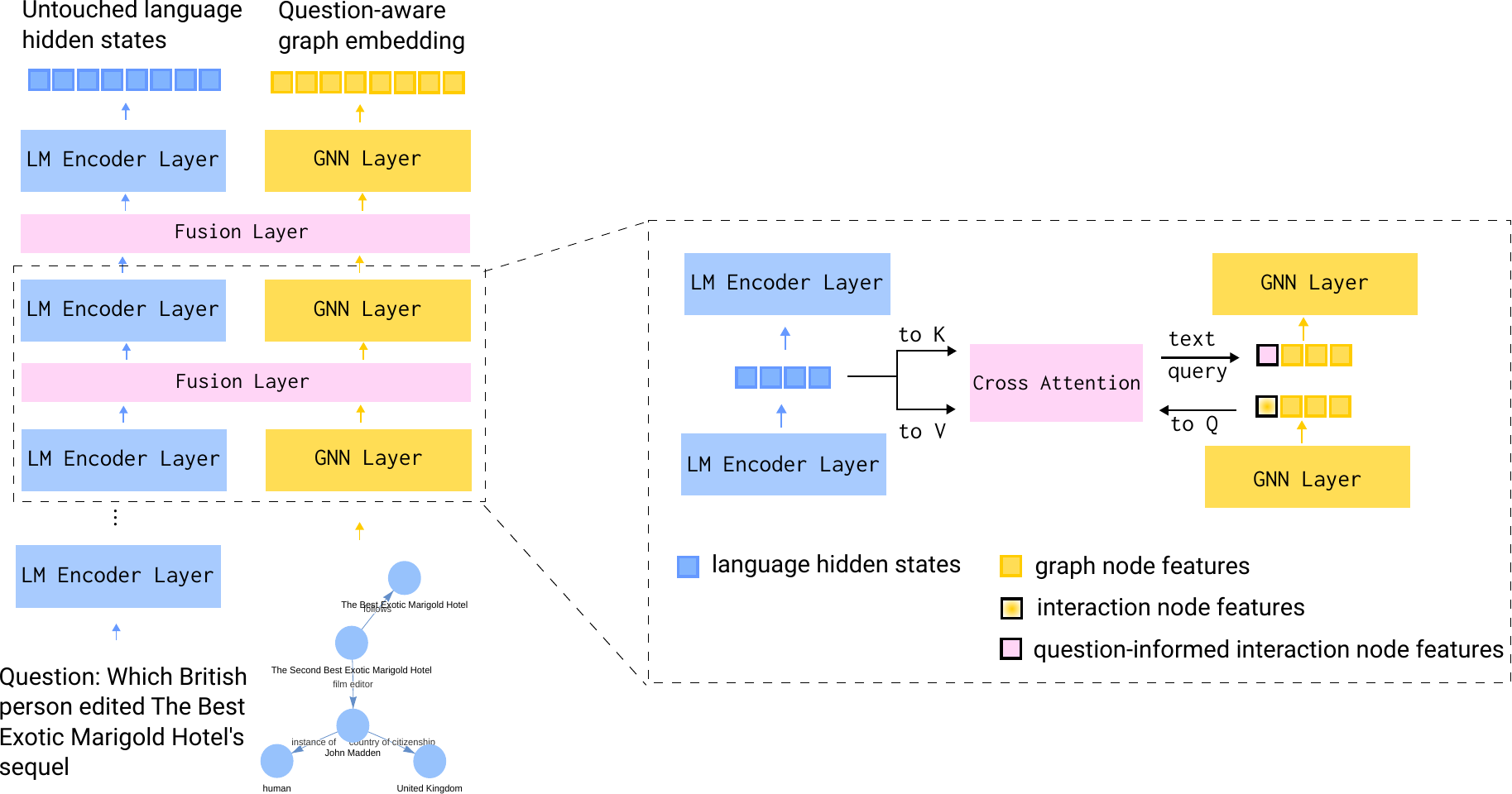}
  \caption{\textbf{The encoder architecture of CrossGNN}. The encoder comprises of a frozen language encoder, a trainable graph encoder, and cross-attention fusion layers. The CrossGNN encoder accepts questions as the language modality input, and a relevant subgraph from Wikidata as the graph modality input, where the node embeddings are initialized with pre-trained KGE. A special interaction node is leveraged to cross-attend question hidden states to incorporate question information into graph modality. The encoded question is left untouched, while the graph is encoded by taking the question into consideration. }
  \label{fig:encoder}
\end{figure*}

The encoder architecture of CrossGNN is shown in Figure \ref{fig:encoder}. It is composed of a frozen language encoder, a graph encoder, and cross-attend fusion layers. The information flow from question to graph encoder is accomplished by creating extra modality interaction nodes in the graph from question hidden states in the last $M$ layers of the $N$ layers of the language encoder, where $M$ equals the number of graph encoder layers, and $N$ equals the number of language encoder layers. The graph embeddings are first initialized with pre-trained knowledge graph embeddings. The modality interaction node is inserted into each of the $M$ graph encoders. It is connected with every other node with a special relation.  In the $k$-th layer of the graph encoder, the interaction node is first updated by a convolutional GNN to gather intra-graph knowledge, then it is used as a query to cross-attend the hidden states of the question from the $N-M+k$-th layers of the language encoder, thereby gathering information from questions. Next, the interaction node updates itself with a feed-forward layer, followed by a residual link. Finally, the question hidden states from the language encoder along with the question-aware graph embedding, including the encoded representation of the interaction node and entity nodes, are passed to the decoder to condition language generation. It is worth noting that the language embeddings are left untouched during the process.

\subsection{Condition Language Generation on the Graph}

\begin{figure}
  \centering
  \includegraphics[width=0.35\textwidth]{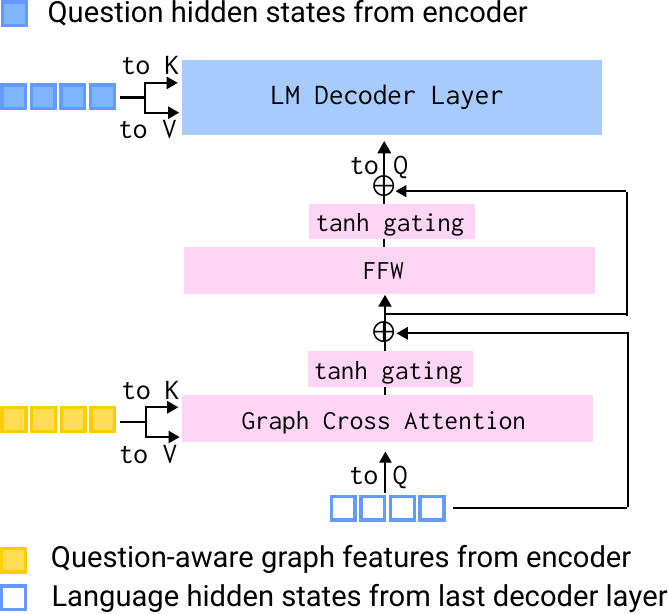}
  \caption{\textbf{A decoder block with injected cross attention from CrossGNN}. Graph cross-attention layers and feed-forward layers are injected into the language decoder at a certain interval to condition text generation on the graph modality. The language decoder is also frozen as the encoder. The injected layers are gated, ensuring that the CrossGNN have the same performance as pre-trained T5 model at initialization. The original cross attention on the encoded question is also left unchanged.}
  \label{fig:decoder}
\end{figure}

We condition a frozen language decoder on question-informed graph representation by inserting gated cross attention into decoding blocks with a certain interval $I$, while the rest of the decoding layers remain unmodified. This approach is inspired by how Flamingo \cite{alayrac2022flamingo} condition language generation with an encoded image or video tokens. The interaction node embedding along with all graph node embeddings are regarded as graph token embeddings which serve as condition signals. At every $I$ decoding layer, the decoder cross attends the graph embeddings to condition the language generation, where the queries are transformed from the language features, and the keys and values are transformed from the graph features. We add a residual link from the input language features to the cross-attended graph features and then pass the aggregated features to the next decoding layer. Following Flamingo, we use a gating mechanism to ensure an unchanged performance with the original language model at initialization. The output of the cross attention is multiplied by a learnable tanh($\alpha$) gate before being added to the residual language features, where $\alpha$ is initialized as zero. In this way, the added cross-attention branch is skipped at initialization, ensuring the outputs of an untrained CrossGNN match that from the pre-trained language model.

%% file: 5_exp.tex
\section{Experiment}

The experiments aims to solve three questions: 1) How much knowledge is already stored in the pretrained language model? 2) Are the paired graphs indeed useful for answer generation? 3) How much improvement can be brought by leveraging the graph modality knowledge? To address these questions, we design three corresponding experiments. First, we examine how much knowledge is carried in the pre-trained language model by fine-tuning a pre-trained T5 model on the questions and answers from GraphextQA alone. This serves as a reference for the rest experiments. Secondly, we validate that the graph does contain useful information for generation by converting the graph to text modality and feeding them as extra context input. To be exact, we finetune a pre-trained T5 model with the questions, verbalized graphs, and answers from GraphextQA. Thirdly, we examine how much information can the language model grasp if we feed in graph knowledge only from graph modality by finetuning CrossGNN on questions, answers, and paired graphs from GraphextQA. This shows the difficulty of leveraging graph modality for text generation.

\begin{table*}
\centering
\begin{tabular}{lllll}
\hline
\textbf{Model} & \makecell{\textbf{graph Modality}} & \textbf{EM} & \textbf{F1} & \textbf{BLEU}\\ 
\hline
T5-base & No graph & 65.73 & 68.26 & 0.5828\\ 
T5-base & Language (verbalized) & 96.29 & 97.64 & 0.8844 \\
CrossGNN & Graph & 68.11 & 70.31 & 0.5946\\
\hline
\end{tabular}
\caption{\label{result}
The evaluation results on GraphextQA of three baselines under different conditions. The first serves as a reference that demonstrates how much information is stored in the pre-trained language model by finetuning a T5 base\footnote{\url{https://huggingface.co/t5-base}} model on GraphextQA. The second T5 base model is trained with verbalized graph as context information in text input, proving that the knowledge encompassed in the graph is useful for text generation. The third one shows the performance of the proposed graph language baseline model CrossGNN, where the answer generation is conditioned on the graph input in graph modality. It demonstrates the difficulty of incorporating graph knowledge into text generation.}
\end{table*}

\subsection{Finetune Language Models with Question-only}

We finetune pre-trained T5-base model on GraphextQA's questions and answers to examine how much knowledge is carried within pre-trained language models. This helps to distinguish the contributions brought by graphs. To conform to the pretraining format of T5, we prepend \textit{Question:} to each question. For questions with multiple answers, we set the first answer and the output target.

\subsection{Finetune Language Models with Questions and Verbalized Graph}

We finetune the pre-trained T5-base model with GraphextQA's questions, answers, and verbalized graphs to prove that the graph contains useful information to answer the questions. First, we verbalize the graphs into texts. Most properties from Wikidata follow a \textit{has-a} semantic, e.g. \textit{[Q345494, P106, Q486748]} expresses that Sakamoto Ryuichi has an occupation of pianist, where Q345494 andDistMult Q486748 are the entity IDs for Sakamoto Ryuichi and pianist, P106 is the property ID for occupation. Therefore, we verbalize each triple in the graph as \textit{\{subject\} has a \{predicate\} \{object\};}, where the identifiers for entities and relations are substituted by their labels. Next, we arrange the input as \textit{question: \{question\}. context: \{verbalized graph\}}, mimicking the preprocessing of SQuAD dataset in the pretraining of the T5 model. The same model architecture and training method as the reference are adopted.

\subsection{Finetune Graph-language Models with Questions and graph}

\subsubsection{Warm-up GNN with Distant Pretraining on Wikipedia Paragraphs}

To warm up the newly added graph-related layers and to better align graph modality for text generation, we pre-train the model with paired graph and text. The pretraining task is to reconstruct the text based on a related graph from Wikidata. This pretraining objective familiarizes the graph encoder with the pre-trained knowledge graph embedding and encourages the model to capture the knowledge encoded in the graph modality.

We acquired such paired graphs and texts by making use of the correspondence between Wikipedia and Wikidata. In many Wikipedia paragraphs, there are some hyperlinks that refer to a mention of another Wikipedia page. Wikimedia maintains a mapping from Wikipedia page to Wikidata entities, which can be found right on the Wikipedia page \footnote{\url{https://en.wikipedia.org/wiki/Wikipedia:Finding\_a\_Wikidata\_ID}}. We further add the links between the entities within a graph by querying Wikidata and adding all links between each of the two entities in a Wikipedia paragraph. To reduce the pretraining burden, we leverage the Wikipedia PageView API to get the most popular Wikipedia items from June 2015 to April 2023. We further remove paragraphs where there are fewer than 4 mentioned entities. This results in a total of  18,810 articles and 144,738 paragraphs with paired graphs for distant pretraining. CrossGNN is pre-trained for 24 epochs.

\subsubsection{Finetune CrossGNN on GraphextQA with}

We use Wikidata embedding pre-trained with TransE \cite{bordes2013transe} from GraphVite \cite{zhu2019graphvite} as pre-trained KGE. It contains pre-trained entity embeddings for 4,818,298 entities. To prepare the graph inputs, we first remove triples without corresponding pre-trained KGE. Then we finetune the warmed-up CrossGNN on GraphextQA, where the training target is to generate answers as natural language based on questions and the paired graph initialized with pre-trained KGE.

\subsection{Results}

Table \ref{result} shows three baseline results on GraphextQA under different conditions. The baseline T5 model trained with no graph involved suggests that pre-trained language models already contain a certain amount of knowledge. The T5-based model finetuned with verbalized graph gains significant improvement over the T5-base baseline without verbalized graph, partly because of the  Being close to 100, it demonstrates that the paired graph in GraphextQA is useful for answer generation in the language modality alone. Furthermore, CrossGNN gains an improvement over the T5-base baseline with 2.38 in EM score and 2.05 in F1 score. But the improvement over the T5 baseline without a graph is very small compared to the results with verbalized graph. On the one hand, it proves CrossGNN's ability to understand and make use of the knowledge from the graph modality, on the other hand, it demonstrates the difficulty for language models to understand graph information, showcasing the difficulty of the proposed dataset and task.

%% file: 6_concl.tex
\section{Conclusion}

We introduce GraphextQA, a multimodal dataset comprising paired questions and graphs, designed to evaluate the integration of cross-modal knowledge from graphs into language generation. We also present CrossGNN, a baseline model that explores the utilization of graph modality for text generation. By comparing evaluation results on language-only models with and without verbalized subgraphs, we prove the usefulness of the paired subgraphs in text generation in the language domain. Moreover, through evaluations conducted on language-only models and the proposed graph-language baseline, CrossGNN exhibits its ability to understand graph modality and leverage it for text generation, evidenced by marginal improvements in EM, F1, and BLEU scores. These results highlight the inherent difficulty in incorporating structured graph modality into the unstructured language modality, emphasizing the need for future research to bridge this gap.

%% file: 7_limit.tex
\section*{Limitations}

\textbf{Question Naturalness}: The majority of questions in the GraphextQA dataset are not natural. Around 87.6\% of the questions are derived from the MCWQ dataset, which employs 29,312 unique question patterns. Conversely, the remaining questions from Lc-QuAD 2.0 are more natural since they are generated by human workers through Amazon Mechanical Turk.

\noindent \textbf{Answer Naturalness}: The answers contained in GraphextQA are text labels for the answer entities, therefore the generation target at training time does not encourage more natural and colloquial answers.

\noindent \textbf{Assumptions and Applicability}: GraphextQA makes strong assumptions about the intended use case.  It assumes that the language model is generative, given the nature of the task, and that the graph information will be leveraged in its native graph modality since the answer are already covered in the labels of the paired subgraph.  These assumptions make GraphextQA not suitable for various existing approaches to knowledge-based question answering (KBQA), including semantic parsing, information retrieval-based methods, and text-only methods.

\noindent \textbf{Limitations of CrossGNN}: One limitation of the proposed CrossGNN model is its dependence on pretrained KGE. The usability of a node in the model depends on the existence of its embedding in the pretrained KGE. However, it is practically impossible to cover the ever-growing entities and relations present in knowledge graphs, as pretrained KGE models must balance coverage and memory consumption. For example, even though the pretrained KGE from graphvite covers 4,818,298 entity embeddings, on average, approximately 1 out of every 4.5 triples from GraphextQA's subgraphs needs to be filtered out due to this limitation.

%% file: 8_ethic.tex
\section*{Ethics Statement}

In terms of ethical considerations regarding the dataset, we implemented OpenAI moderation APIs\footnote{\url{https://platform.openai.com/docs/api-reference/moderations}} to screen for potentially harmful questions, including those involving violence, sexual content, or hate speech. The results revealed that no questions were flagged as containing harmful content.

Regarding potential ethical concerns with the CrossGNN model, it utilizes both pretrained knowledge embedded in the model and the graph modality for text generation. Consequently, CrossGNN has the potential to manifest biases and incorporate toxic information present within knowledge graphs and pretrained language models.
